\documentclass[letterpaper, 10 pt, conference]{ieeeconf}

\IEEEoverridecommandlockouts
\usepackage[T1]{fontenc}
\usepackage[utf8]{inputenc}

\usepackage{graphicx}
\graphicspath{{images/}{figures/}}

\usepackage{amsmath}
\usepackage{amssymb}
\usepackage{nicefrac}
\usepackage{siunitx}

\usepackage{booktabs}
\usepackage{multirow}
\usepackage{float}
\usepackage{xcolor}

\let\labelindent\relax
\usepackage{enumitem}
\setlist[itemize]{align=parleft,left=0pt..1em}

\usepackage[font=small]{caption}
\usepackage{subcaption}

\usepackage{algorithm}
\usepackage{algorithmic}

\usepackage{listings}

\usepackage{acro}

\usepackage{hyperref}

\usepackage{flushend}

\title{\LARGE \bf Edge Radar Material Classification Under Geometry Shifts}
\author{}
\author{Jannik Hohmann, Dong Wang, and Andreas N{\"u}chter
\thanks{This work was in parts supported by the Federal Ministry for Economic Affairs and Climate Action (BMWK) on the basis of a decision by the German Bundestag under the grant number KK5150106RL4.}%
\thanks{Jannik Hohmann, Dong Wang {\tt\small (dong.wang@uni-\allowbreak wuerzburg.de)} and Andreas N{\"u}chter are with the Department of Informatics XVII (Robotics), Julius-Maximilians-University W{\"u}rzburg, 97074 W{\"u}rzburg, Germany. Andreas N{\"u}chter is also with the Zentrum f{\"u}r Telematik e.V., W{\"u}rzburg and currently International Visiting Chair at U2IS, ENSTA, Institut Polytechnique de Paris, France.}%
}

\begin{document}

\maketitle
\thispagestyle{empty}
\pagestyle{empty}

\begin{abstract}
Material awareness can improve robotic navigation and interaction, particularly in conditions where cameras and LiDAR degrade.
We present a lightweight mmWave radar material classification pipeline designed for ultra-low-power edge devices (TI IWRL6432), using compact range-bin intensity descriptors and a Multilayer Perceptron (MLP) for real-time inference.
While the classifier reaches a macro-F1 of 94.2\% under the nominal training geometry, we observe a pronounced performance drop under realistic geometry shifts, including sensor height changes and small tilt angles.
These perturbations induce systematic intensity scaling and angle-dependent radar cross section (RCS) effects, pushing features out of distribution and reducing macro-F1 to around 68.5\%.
We analyze these failure modes and outline practical directions for improving robustness with normalization, geometry augmentation, and motion-aware features.
\end{abstract}

\section{INTRODUCTION}
Robots deployed for long periods in the wild must maintain reliable perception under changing environments and gradual sensor drift.
Material classification is a deployment-sensitive capability: it can inform traversability and interaction decisions, yet sensing geometry (height, tilt, and mounting tolerances) inevitably varies over time.
Radar is attractive for such settings because it is robust to lighting and many adverse conditions, and can operate in privacy-sensitive environments~\cite{TI_FMCW_radar}.

Practical radar material classification on embedded devices faces two coupled challenges.
First, edge hardware imposes tight compute and power budgets, favoring compact feature representations and small networks~\cite{TI_Machine_Learning_mmWave_Radar}.
Second, intensity-based radar signatures are strongly geometry-dependent: received power varies with range and changes with angle-dependent RCS, so small mounting variations can yield out-of-distribution (OOD) inputs at deployment.

This paper reports a baseline implementation and an evaluation centered on geometry shifts.
We implement a lightweight pipeline on the TI IWRL6432 platform that extracts a 12-dimensional range-bin intensity descriptor and performs real-time inference with an MLP on-device.
Under the nominal geometry, the model achieves a macro-F1 of 94.2\%.
However, performance degrades substantially under realistic geometric perturbations (height and tilt), revealing a robustness gap.
We analyze the dominant failure modes and provide a roadmap toward more geometry-aware and deployment-robust radar material classification that remains compatible with embedded constraints.

This paper makes the following contributions:
\begin{itemize}
\item We present a practical edge baseline for radar material classification on TI IWRL6432, combining a compact intensity descriptor with an MLP for real-time on-device inference.
\item We quantify robustness under geometry shifts (height and tilt) and under session-level variations using macro-F1 and confidence statistics.
\item We identify failure modes caused by range-dependent intensity scaling and angle-dependent RCS changes, and outline mitigation strategies (normalization, augmentation, and motion-aware features).
\end{itemize}

\section{RELATED WORK}
FMCW mmWave radar has become a key modality for robotic perception due to its robustness in degraded visibility and adverse conditions~\cite{TI_FMCW_radar}.
Recent work emphasizes resource-efficient on-device processing, enabling ``machine learning on the edge'' for low-power radar platforms~\cite{TI_Machine_Learning_mmWave_Radar}~\cite{ti_iwrl6432_ds2025}~\cite{ti_iwrl6432boost2025}.
Such constraints often motivate compact feature representations and lightweight models (e.g., small MLPs) when memory and latency budgets are strict~\cite{Goodfellow-et-al-2016}.

Radar-based material (or surface) classification leverages differences in reflectivity, dielectric properties, and scattering behavior, but signatures can be sensitive to geometry, multipath, and sensor-dependent factors, especially when relying on absolute intensity cues~\cite{Physik}~\cite{radar_signature}.
Beyond intensity-only descriptors, learning-based approaches increasingly use richer inputs such as IQ signals or radar images.
For example, SMCNet demonstrates supervised surface material classification using mmWave radar IQ signals with complex-valued CNNs and explicitly studies distance generalization~\cite{haegele2025smcnet}.
Material-ID explores mmWave-based material identification with broader system considerations and data-driven modeling~\cite{chen2025materialid}.
Earlier work such as RaCaNet shows that radar imaging with deep CNNs can achieve high classification accuracy in controlled settings for 60\,GHz radar material classification~\cite{weiss2019racanet}.

Related RF sensing literature (outside strict mmWave radar pipelines) further highlights that material recognition can be feasible from RF propagation effects, but also prone to distribution shifts.
RadarCat explores radar categorization for interactive sensing, illustrating the utility of radar signatures for recognition tasks under practical constraints~\cite{yeo2016radarcat}.
Wi-Fi based material sensing similarly studies how channel responses can encode material properties, emphasizing feasibility but also sensitivity to environment and setup~\cite{zhang2019wifi_material}.
In industrially motivated settings, 60\,GHz radar has also been explored for material sorting/classification in recycling-oriented scenarios, showing strong accuracy under constrained setups~\cite{albing2023recycling60ghz}.
More recently, feature-selection-based ML pipelines have been investigated for 60\,GHz FMCW radar material classification, again underlining that the feature/geometry interplay is critical for performance~\cite{nongpromma2025ecticon}.

OOD robustness and long-term reliability are widely recognized as central challenges for deployment-grade perception~\cite{Schwaiger2020OOD}~\cite{Henne2021ReliablePerception}.
In addition, run-time monitoring is increasingly viewed as a practical safety layer to detect or anticipate failures of learning-based perception components during operation~\cite{rahman2021runtime_monitoring}.
Motivated by these insights, our work quantifies how small, realistic geometric changes (height and tilt) can break an edge-deployed intensity-based mmWave material classifier, and highlights mitigation directions compatible with embedded constraints.

\section{METHODOLOGY}
The pipeline follows a deterministic signal-processing chain designed for high-frequency real-time execution.
Raw radar measurements are transformed into compact feature vectors, which are then classified by a lightweight neural network.

\subsection{Data Acquisition and Hardware Configuration}
Data collection was performed using the Texas Instruments IWRL6432BOOST, an integrated single-chip FMCW mmWave radar sensor operating in the 60\,GHz band~\cite{Board_Manual}.
The sensor was mounted in a nadir-looking geometry at a constant height of $H=45$\,cm with a nominal downward inclination of $\theta=90^\circ$ relative to planar material samples.
We considered five material classes: iron, aluminum, plexiglass, wood, and limestone.
To reduce stochastic temporal noise, intensity measurements were integrated over a fixed number of frames per recorded sequence (identical setting across all classes).

\begin{figure}[htbp]
    \centering
    \includegraphics[width=0.85\linewidth]{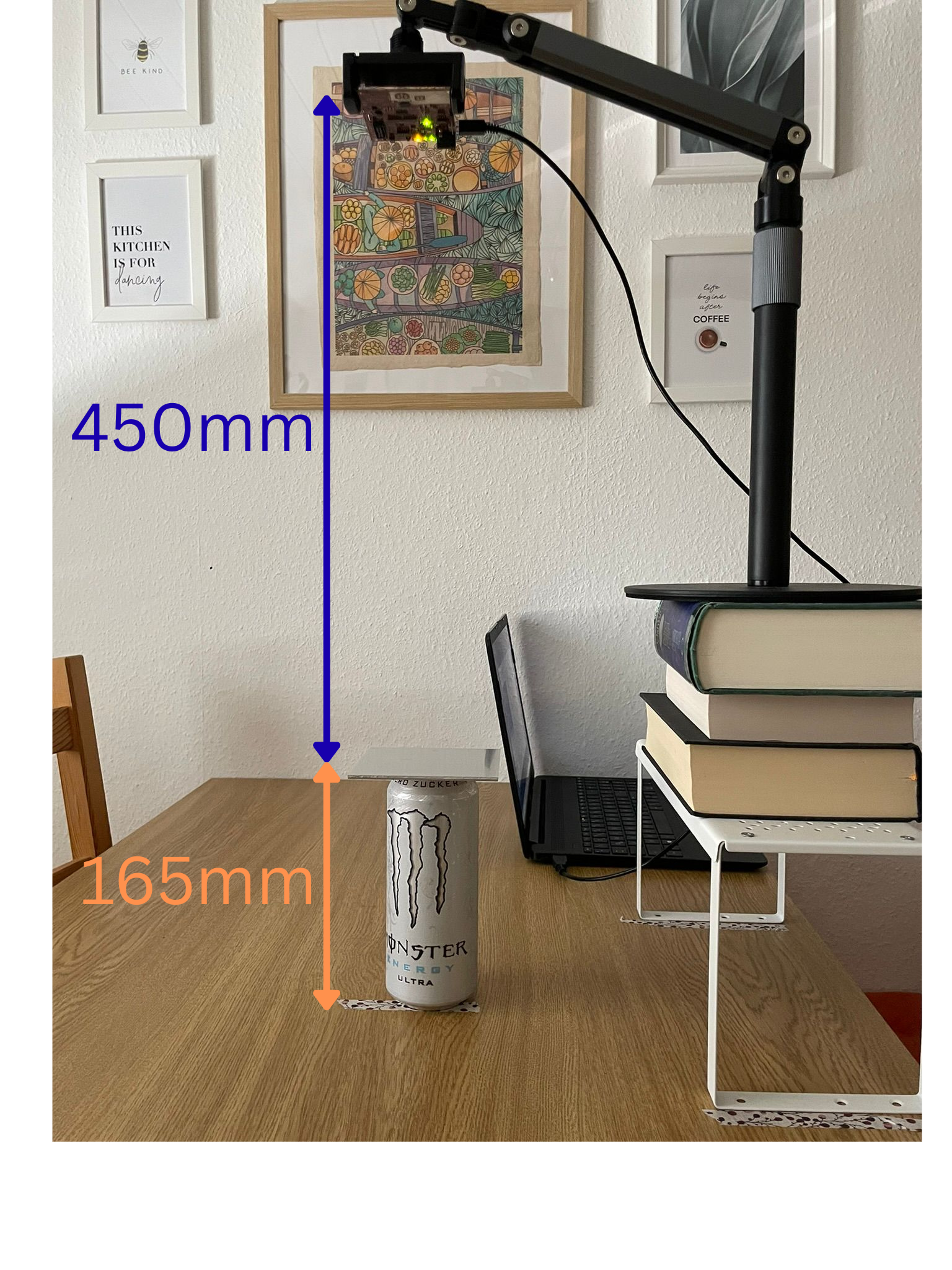}
    \caption{Default experimental setup used for baseline data acquisition. The radar is mounted in a nadir-facing configuration at $H=45$\,cm and $\theta=90^\circ$ relative to the material surface.}
    \label{fig:default_setup}
\end{figure}

Fig.~\ref{fig:default_setup} illustrates the default acquisition geometry used as the in-distribution reference domain (\textit{Known Data}).
All robustness experiments modify one factor relative to this reference setup: sensor height, inclination angle, or recording session/sample identity.

\subsection{Feature Extraction}
The extraction process converts raw Analog-to-Digital Converter (ADC) data into range-intensity profiles.
To comply with edge constraints, we use a low-dimensional descriptor based on range-dependent magnitudes instead of full range-Doppler tensors.

\subsubsection{Range Bin Selection}
We determine an informative range-bin window by analyzing energy concentration on the training set and selecting the bins that consistently cover the dominant return from the planar target at nominal geometry.
In our setup, relevant material-specific information concentrates within Range Bins 6--17.
Bins outside this interval predominantly contain background clutter or noise-floor components and are excluded to reduce compute and improve stability.

\subsubsection{Intensity Descriptor}
For each frame, the magnitudes of the 12 selected range bins are concatenated into a 12-dimensional feature vector.
This descriptor acts as direct input to the classifier and captures material-dependent backscattering characteristics shaped by dielectric properties and scattering coefficients~\cite{Physik}.

\subsection{Model Architecture and Edge Deployment}
We implement a Multilayer Perceptron (MLP) to satisfy latency and memory constraints on the IWRL6432.
The network consists of:
\begin{itemize}
    \item \textbf{Input layer:} 12 nodes (selected range bins).
    \item \textbf{Hidden layers:} two fully connected layers with Batch Normalization~\cite{ioffe2015batch} and ReLU activations~\cite{Nair2010RectifiedLU}.
    \item \textbf{Output layer:} 5 nodes with Softmax to obtain class probabilities.
\end{itemize}

The model is trained using the Adam optimizer with cross-entropy loss.
After training, it is deployed on-device via the IWRL6432 software stack, achieving an average on-device inference latency of approximately 20\,$\mu$s (MLP forward pass), enabling execution within the sensor processing cycle without external compute.

\section{EXPERIMENTAL EVALUATION}
We first quantify performance under nominal (training) geometry and then probe generalization under realistic geometric perturbations.
We report macro-averaged F1-score and analyze Softmax confidence distributions.

\subsection{Baseline Performance under Nominal Geometry}
The baseline evaluation uses the nominal configuration ($H=45$\,cm, $\theta=90^\circ$).

\subsubsection{Classification Accuracy and Confusion Analysis}
\begin{figure}[htbp]
    \centering
    \includegraphics[width=0.85\linewidth]{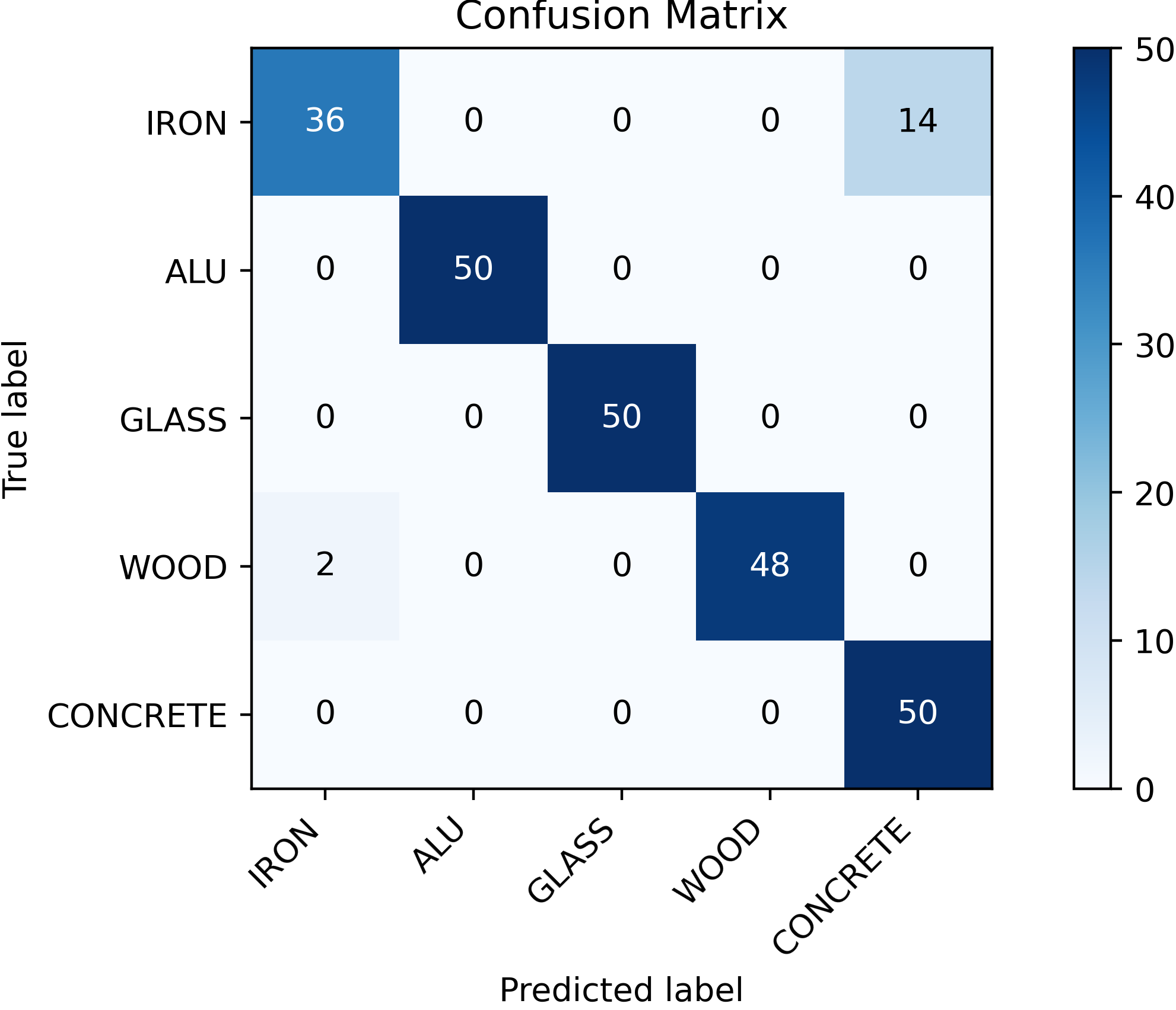}
    \caption{Confusion matrix of the MLP classifier under nominal conditions ($H=45$\,cm, $\theta=90^\circ$). Diagonal entries indicate per-class recall.}
    \label{fig:confusion_baseline}
\end{figure}

The classifier reaches a macro-F1 score of 94.2\% on the validation split.
The confusion matrix in Fig.~\ref{fig:confusion_baseline} shows strong separability of metallic vs.\ non-metallic classes.
Residual confusion is mainly between iron and aluminum, which exhibit similar reflectivity in the 60\,GHz band.

\subsubsection{Confidence Distribution}
\begin{figure}[htbp]
    \centering
    \includegraphics[width=0.85\linewidth]{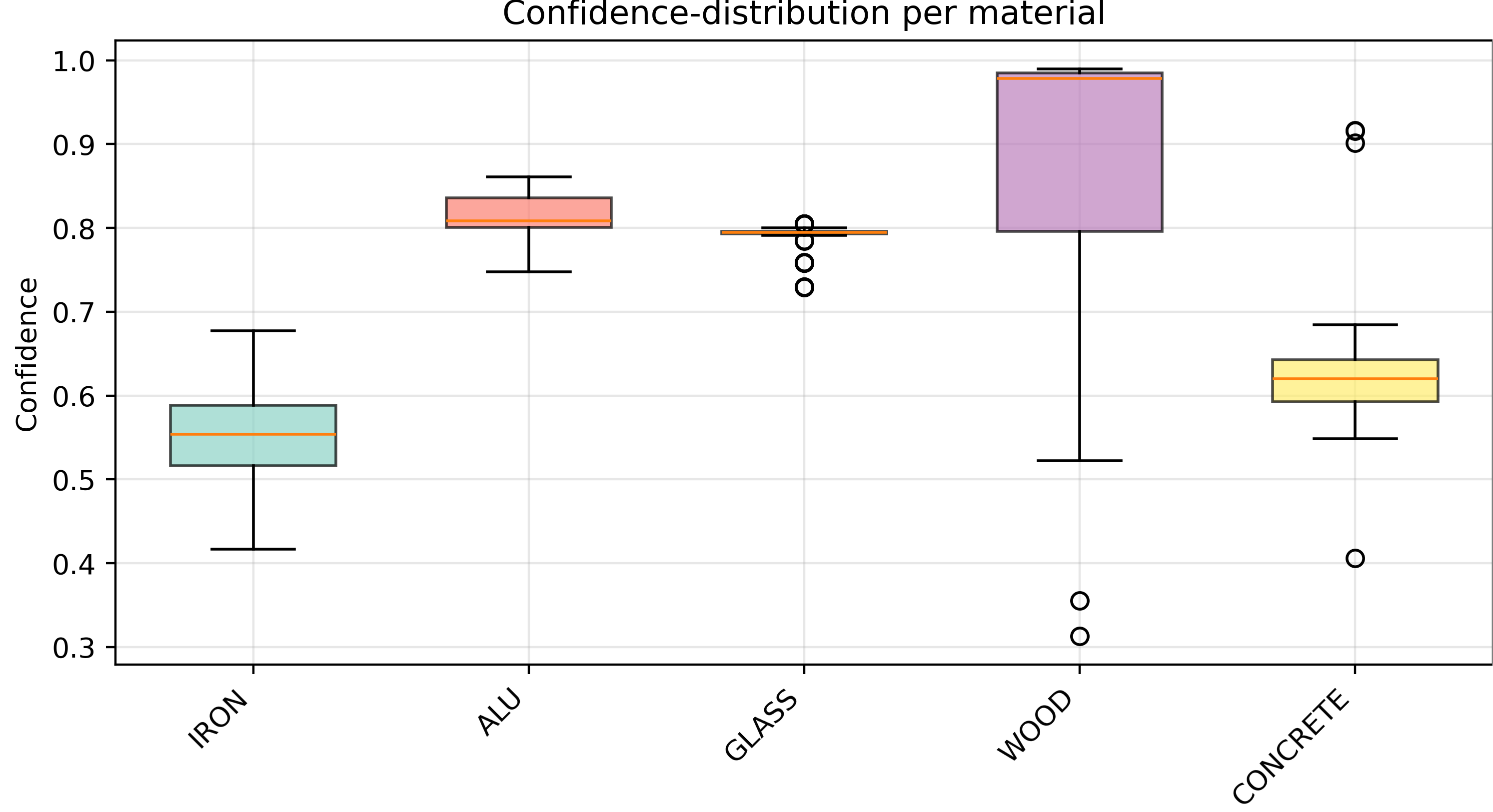}
    \caption{Confidence (maximum Softmax probability) distribution under nominal geometry.}
    \label{fig:confidence_baseline}
\end{figure}

For the nominal dataset, correct predictions typically yield high Softmax maxima.
Fig.~\ref{fig:confidence_baseline} shows that confidence values for correct classifications often exceed $p=0.9$, while misclassifications correlate with lower confidence, consistent with proximity to learned decision boundaries.

\subsection{Performance under Geometry Shifts}
We evaluate two geometry perturbations relative to the nominal setup: height change ($H \in \{35,55\}$\,cm) and tilt change ($\theta = 90^\circ \pm 10^\circ$).

\subsubsection{Impact of Sensor Height Displacement}
\begin{figure}[htbp]
    \centering
    \includegraphics[width=0.85\linewidth]{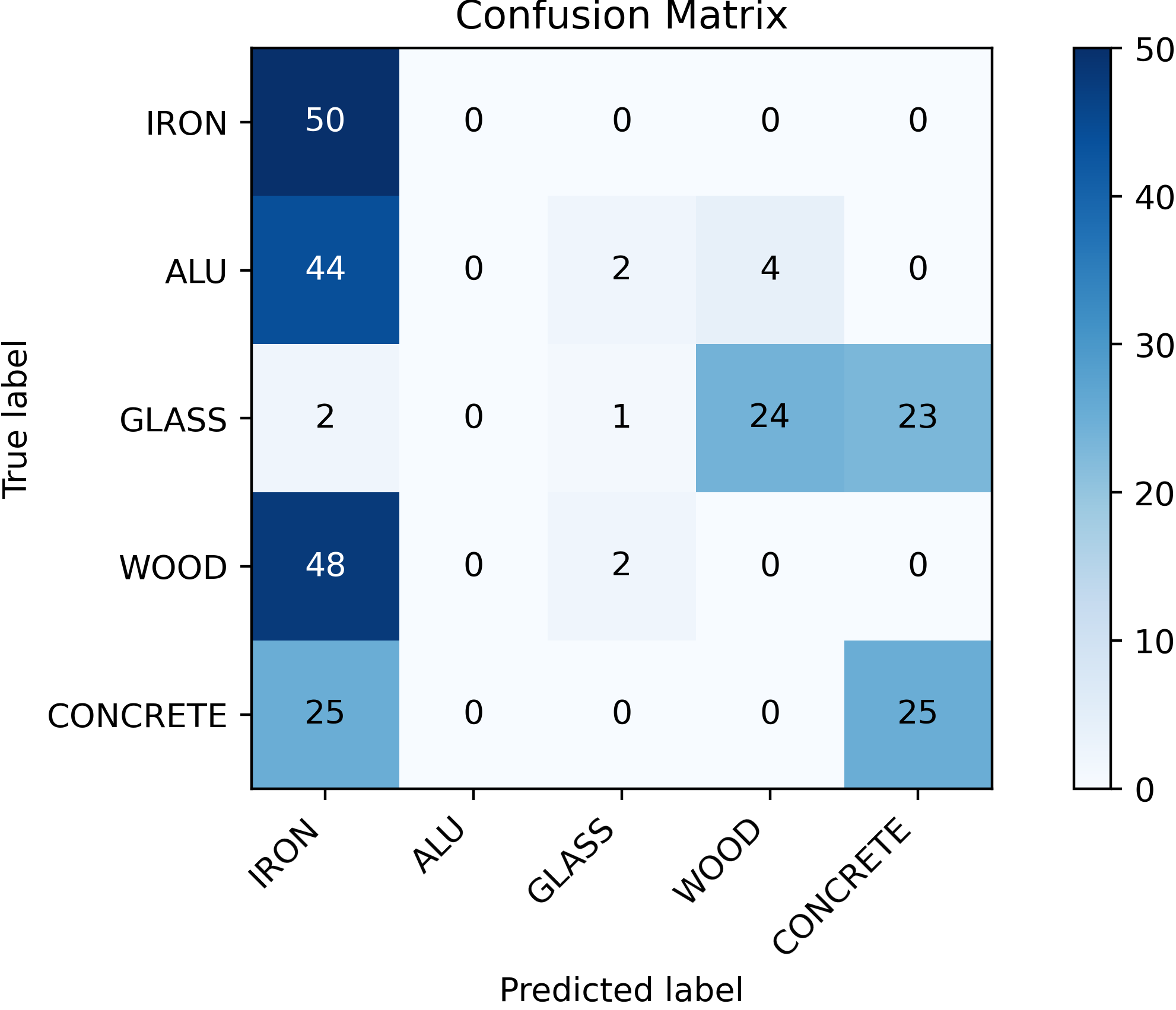}
    \caption{Confusion matrix under height variations (35\,cm and 55\,cm). Off-diagonal density increases due to distance-dependent power changes.}
    \label{fig:confusion_height}
\end{figure}

We vary sensor height between 35\,cm and 55\,cm.
Performance degrades as range deviates from the training height, consistent with the strong range dependence of received power.
Because the MLP operates on absolute intensities within fixed range bins, feature vectors drift relative to learned decision boundaries.
Fig.~\ref{fig:confusion_height} shows increased off-diagonal mass, particularly when increased distance reduces the apparent intensity of reflective materials.

\subsubsection{Impact of Inclination Angle and RCS}
\begin{figure}[htbp]
    \centering
    \includegraphics[width=0.85\linewidth]{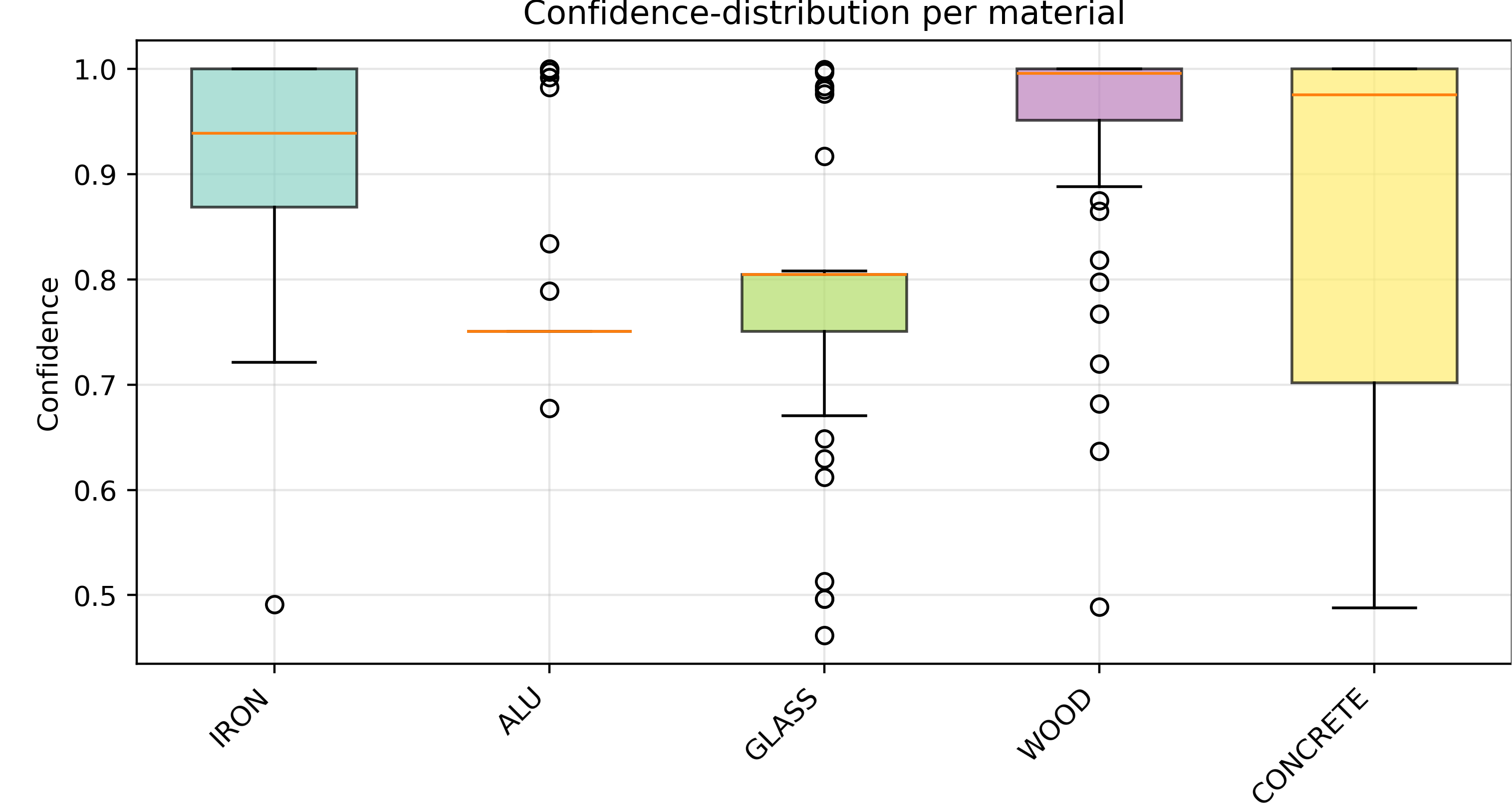}
    \caption{Confidence distribution under geometry shifts (tilt). Increased low-confidence predictions indicate reduced certainty under OOD geometry.}
    \label{fig:confidence_unknown}
\end{figure}

We tilt the sensor by $\pm 10^\circ$ from nadir.
This perturbs the effective RCS of planar targets; for specular reflectors (iron, aluminum), small angular changes can redirect energy away from the monostatic receiver, reducing measured intensity.
We observe a macro-F1 decrease to approximately 68.5\%.
The confidence distribution in Fig.~\ref{fig:confidence_unknown} shifts toward lower values, indicating elevated uncertainty under OOD geometry.

\subsection{Performance under Similar Geometry with Independent Recordings}
To separate geometry shift from session-level variability, we evaluate a ``similar but unknown'' dataset recorded under nominal geometry ($H=45$\,cm, $\theta=90^\circ$) but with independent recording sessions and different samples.

\subsubsection{Generalization to Material Classes}
\begin{figure}[htbp]
    \centering
    \includegraphics[width=0.85\linewidth]{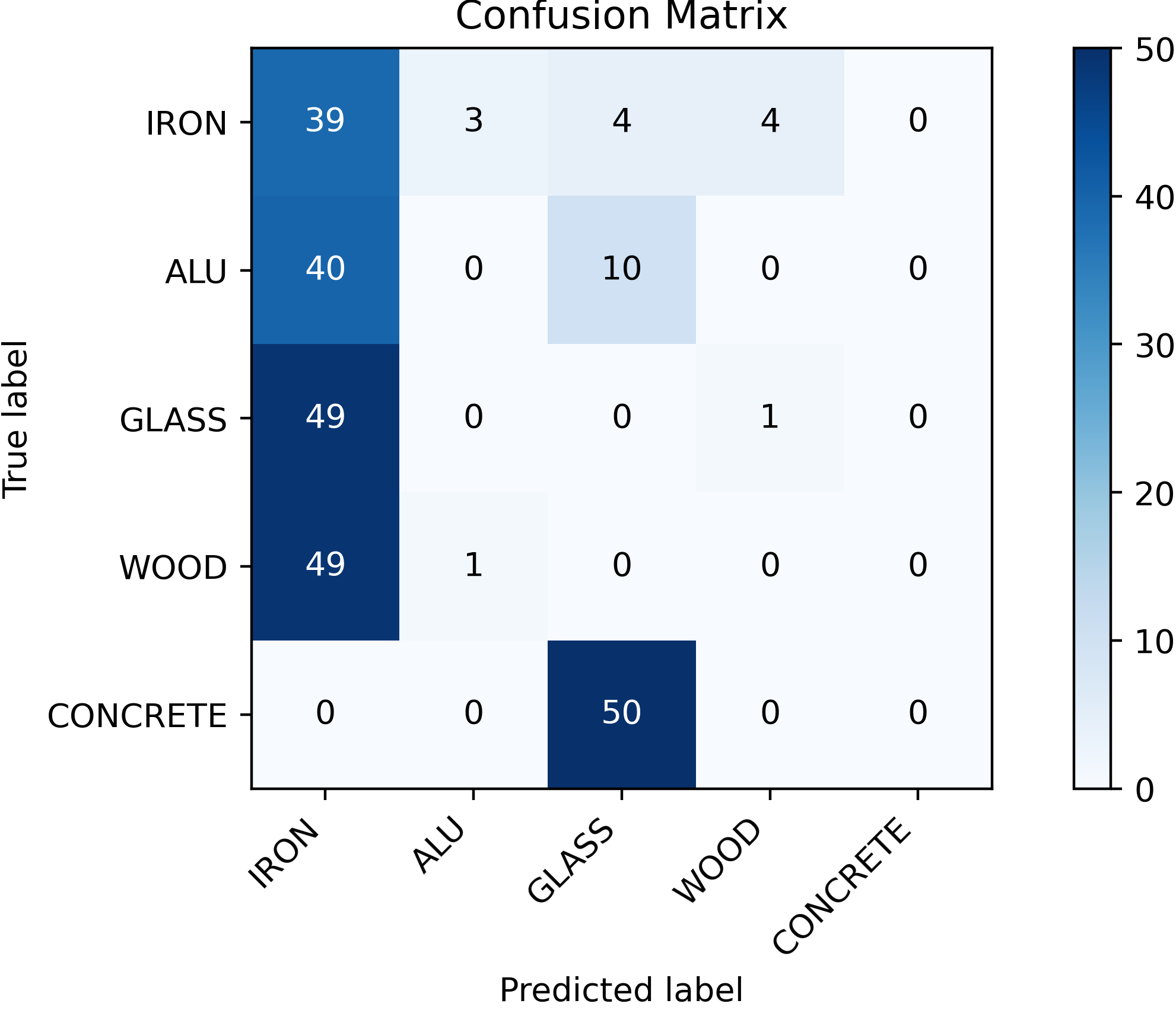}
    \caption{Confusion matrix for ``similar but unknown'' data at nominal geometry ($H=45$\,cm, $\theta=90^\circ$). Increased confusion indicates session-level sensitivity despite unchanged geometry.}
    \label{fig:confusion_similar}
\end{figure}

Compared to the baseline validation, we observe a notable performance drop and increased confusion (Fig.~\ref{fig:confusion_similar}).
This suggests sensitivity to session-specific intensity fluctuations and partial overfitting to the training recordings' noise characteristics, particularly for low-contrast material pairs.

\subsubsection{Comparative Confidence Analysis}
\begin{figure}[htbp]
    \centering
    \includegraphics[width=0.85\linewidth]{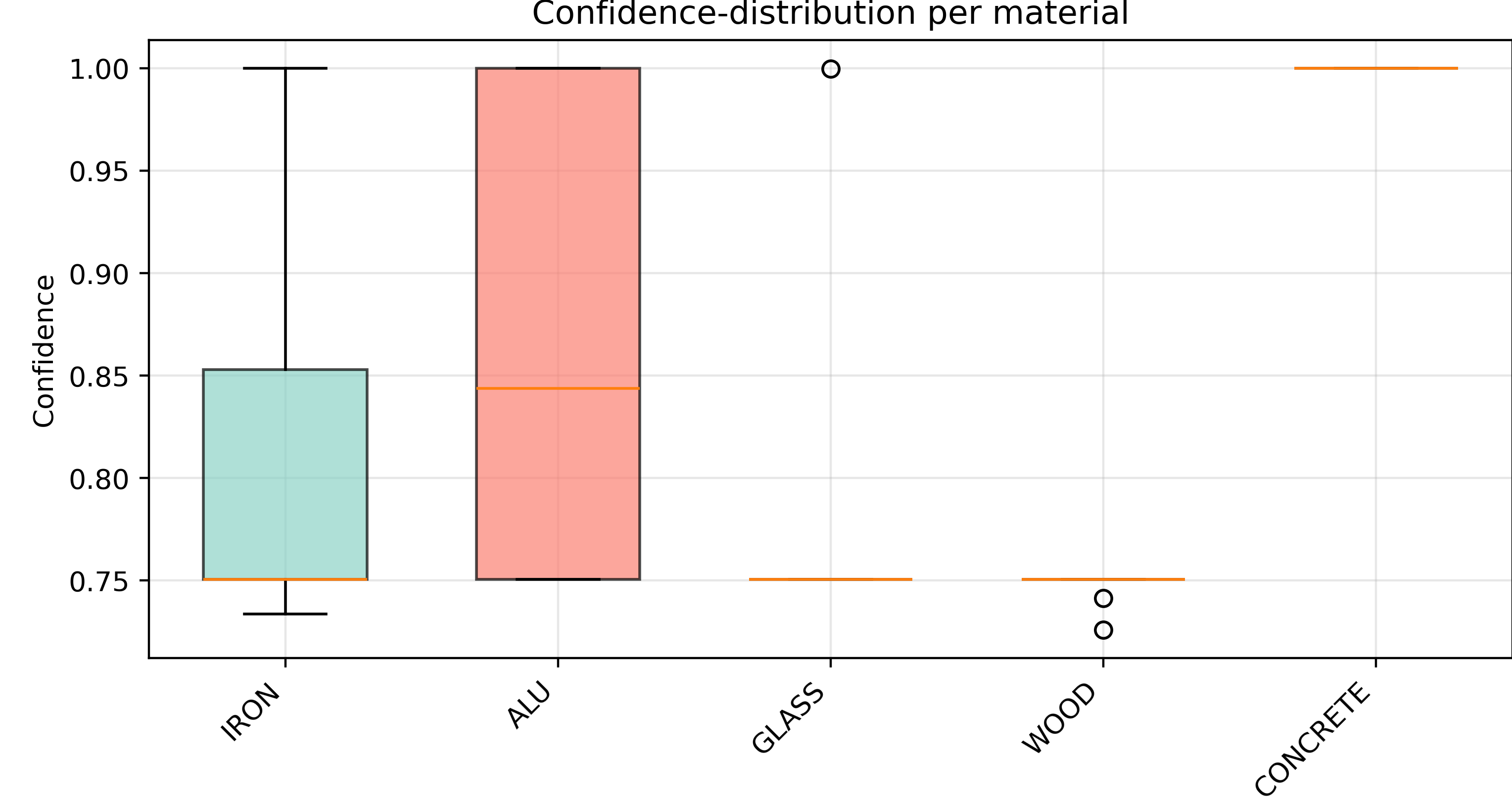}
    \caption{Confidence distribution for ``similar but unknown'' data at nominal geometry. The probability mass shifts toward lower confidence compared to the baseline.}
    \label{fig:confidence_similar}
\end{figure}

Fig.~\ref{fig:confidence_similar} shows a shift toward lower confidence compared to the baseline, consistent with feature vectors moving closer to decision boundaries due to session-level intensity changes and SNR fluctuations.

\section{CONCLUSION AND DISCUSSION}
We presented a lightweight mmWave radar material classification pipeline for ultra-low-power edge deployment on the TI IWRL6432BOOST~\cite{Board_Manual}.
Using a 12-dimensional range-bin intensity descriptor and a compact MLP, the system achieves strong in-distribution performance (macro-F1 94.2\%) with inference latency around 20\,$\mu$s, making it suitable for embedded operation~\cite{TI_Machine_Learning_mmWave_Radar}.

\subsection{Failure Modes under Geometry Shifts}
A central finding is that absolute intensity features are not geometry-invariant.
Height changes induce systematic scaling of received power, while small tilt angles alter effective RCS and can strongly attenuate specular returns~\cite{radar_signature}.
Both effects yield OOD feature drift and large performance degradation (macro-F1 down to $\sim$68.5\% under tilt).
Furthermore, even under unchanged geometry, independent recordings exhibit session-level variability that reduces confidence and accuracy, indicating sensitivity to temporal noise patterns.

\subsection{Roadmap Toward Robust Deployment}
Our results suggest three practical directions that remain compatible with embedded constraints:
\begin{itemize}
    \item \textbf{Range-aware normalization:} compensate distance-dependent scaling (e.g., calibration curves or physics-guided normalization).
    \item \textbf{Geometry augmentation:} train with height/tilt perturbations to enlarge the support of the training distribution.
    \item \textbf{Motion-aware features:} explore Doppler or micro-motion signatures to reduce reliance on absolute intensity cues~\cite{doppler}.
\end{itemize}
In addition, simple uncertainty monitoring using confidence statistics can provide a lightweight trigger for re-calibration or fallback behaviors in long-term operation.

\subsection{Future Work}
Beyond normalization and augmentation, future work should investigate lightweight architectures with improved inductive bias (e.g., small CNNs) while respecting memory and latency constraints.
Disciplined hyper-parameter selection (learning rate schedules, weight decay) may further stabilize training across recording sessions~\cite{smith2018disciplinedapproachneuralnetwork}.

\paragraph{Limitations}
Our study focuses on planar samples and a single radar platform/configuration; broader target geometries, additional sensors, and outdoor conditions are left for future work.

\bibliographystyle{IEEEtran}
\bibliography{bibliography.bib}

\end{document}